\documentclass[letter, 10 pt, conference]{ieeeconf}  
\IEEEoverridecommandlockouts                              

\usepackage{hyperref}
\usepackage{placeins}
\usepackage{amssymb}
\usepackage{pifont}
\newcommand{\xmark}{\ding{55}}%
\usepackage{amsmath}
\usepackage{bm}

\usepackage{units}
\usepackage{multirow}

\usepackage{tikz}
\usetikzlibrary{external,arrows,shapes,automata,backgrounds,petri,fit,arrows.meta,positioning,calc,matrix,chains,scopes}
\usepackage[skins]{tcolorbox}
\usepackage{pgfplotstable}
\usepackage{pgfplots}
\pgfplotsset{compat=newest}
\pgfplotsset{minor grid style={dashed,white!80!black}}
\pgfplotsset{major grid style={white!80!black}}

\definecolor{ownBlue}{rgb}{0.0705882,0.694118,0.839216}
\definecolor{ownGreen}{rgb}{0.631373,0.784314,0.25098}
\definecolor{ownOrange}{rgb}{0.996078,0.741176,0.239216}
\definecolor{ownYellow}{rgb}{0.960784,0.898039,0.14902}
\definecolor{ownRed}{rgb}{0.990784,0.298039,0.14902}

\newcommand\copyrighttext{%
  \footnotesize \textcopyright 2021 IEEE. Personal use of this material is permitted. Permission from IEEE must be obtained for all other uses, in any current or future media, including reprinting/republishing this material for advertising or promotional purposes, creating new collective works, for resale or redistribution to servers or lists, or reuse of any copyrighted component of this work in other works.
  DOI: \href{https://ieeexplore-ieee-org.thi.idm.oclc.org/document/9575730}{10.1109/IV48863.2021.9575730}}
\newcommand\copyrightnotice{%
\begin{tikzpicture}[remember picture,overlay]
\node[anchor=south,yshift=10pt] at (current page.south) {\fbox{\parbox{\dimexpr\textwidth-\fboxsep-\fboxrule\relax}{\copyrighttext}}};
\end{tikzpicture}%
}

\title{\LARGE \bf
Novelty Detection and Analysis of Traffic Scenario Infrastructures in the Latent Space of a Vision Transformer-Based Triplet Autoencoder}
\author{Jonas Wurst$^{1}$, Lakshman Balasubramanian$^{1}$, Michael Botsch$^{1}$ and Wolfgang Utschick$^{2}$
\thanks{$^{1}$CARISSMA, Technische Hochschule Ingolstadt, 85049 Ingolstadt, Germany
        {\tt\small\{firstname.lastname\}@thi.de}}%
\thanks{$^{2}$Technical University of Munich, 80333 Munich, Germany        {\tt\small utschick@tum.de}}%
}

\begin{document}
\bstctlcite{IEEEexample:BSTcontrol}

\maketitle
\copyrightnotice
\thispagestyle{empty}
\pagestyle{empty}
\begin{abstract}
	Detecting unknown and untested scenarios is crucial for scenario-based testing. Scenario-based testing is considered to be a possible approach to validate autonomous vehicles. A traffic scenario consists of multiple components, with infrastructure being one of it. In this work, a method to detect novel traffic scenarios based on their infrastructure images is presented. An autoencoder triplet network provides latent representations for infrastructure images which are used for outlier detection. The triplet training of the network is based on the connectivity graphs of the infrastructure. By using the proposed architecture, expert-knowledge is used to shape the latent space such that it incorporates a pre-defined similarity in the neighborhood relationships of an autoencoder. An ablation study on the architecture is highlighting the importance of the triplet autoencoder combination. The best performing architecture is based on vision transformers, a convolution-free attention-based network. The presented method outperforms other state-of-the-art outlier detection approaches.
\end{abstract}
\section{INTRODUCTION}\label{sec:intro}
A simple statistical proof of an \textit{Autonomous Vehicle's} (AV's) safety is infeasible. Approaches like scenario-based testing are used for validation. There, the testing of an AV is focused on relevant scenarios, instead of exclusively driving randomly millions of kilometers. Identifying representative scenarios is required for this approach \cite{Junietz2018a}. The scenarios can be constructed by expert-knowledge or from real world driving data. A key aspect of the latter strategy is to identify new scenarios which have not been tested yet. The two commonly used approaches to detect unknown scenarios are cluster assignment quality or outlier detection. The latter is realized in this work.

A traffic scenario is described by multiple aspects, for example the dynamics and the environment \cite{PEGASUS}. Publications often focus on the dynamics to identify representative and novel scenarios (e.\,g. \cite{Wang2020a}, \cite{Harmening2020a}, \cite{Demetriou2020a}, \cite{Hauer2020a}). Besides dynamics, another crucial component of a scenario is the infrastructure. Here, birds-eye view images, representing the infrastructure, are used to detect novel traffic scenarios.

In this work, a method to detect unknown and potentially untested infrastructures is presented. For this purpose, the infrastructure images are projected into a latent space using a deep learning pipeline. As will be shown, a high performance increase is achieved when using the latent space instead of the input space for novelty detection. In the latent space, simple outlier detection methods can be used to identify novel infrastructures, which indicates that the method is able to generate strong representations. In order to create this latent space, an autoencoder architecture utilizing metric learning via triplet loss is used. The triplet mining is based on the connectivity of the infrastructures. Through the combination of the autoencoder scheme and the triplet learning, expert-knowledge is used for shaping the latent space. 

Extensive evaluation of the presented method is performed. For the encoder, state-of-the-art networks such as \textit{Vision Transformers} (ViTs) \cite{Dosovitskiy2021a} and ResNet-18 \cite{He2016a} are evaluated. Experiments demonstrate the influence of the triplet loss as well as the autoencoder scheme. The resulting architecture combinations are outperforming the alternative methods. An implementation of the architecture is made publicly available\footnote{\url{https://github.com/JWTHI/ViTAL-SCENE}}.

The contributions of this work can be summarized as:
\begin{enumerate}
	\item A new method for novelty detection of infrastructure images in the latent space of triplet loss-based autoencoder networks is presented.
	\item Automated triplet mining of road infrastructures without manual labeling is introduced.
	\item The performance is evaluated against various state-of-the-art methods and shows significant improvements.
\end{enumerate}

This work is organized as follows. In Sec. \ref{sec:relWork} the related work in the field of novel traffic scenario detection and clustering are discussed and compared to this paper. The method, consisting of the data generation, triplet mining, triplet network and the outlier detection is introduced in Sec. \ref{sec:method}. Sec. \ref{sec:exps} presents the experimental results, when applying the proposed method to a road infrastructure data set. Finally, the work is concluded in Sec. \ref{sec:conc}.

\section{RELATED WORK}\label{sec:relWork}
\subsection{Traffic Scenario Identification}
The validation of AVs through scenario-based testing is considered to be one of the possible approaches to prove their safety \cite{Junietz2018a}. In the survey \cite{Riedmaier2020a}, various works utilizing the scenario-based validation approach are summarized. The survey also lists works trying to identify and define relevant scenarios, of which some are based on machine learning.

Several works use clustering for novelty detection. The underlying assumption is that scenarios which do not belong to a certain cluster, can be treated as novel scenarios. Since in this work a learned representation for the infrastructural part of a scenario is introduced, first the works are grouped with respect to their used scenario similarity or the used scenario representation. In \cite{Kruber2018a} and \cite{Kruber2019b} a similarity measure based on an unsupervised random forest is used. More similar to this work are \cite{Wang2020a} (LSTM+CNN), \cite{Harmening2020a} (SeqDSPN), \cite{Demetriou2020a} (RC-GAN), where the latent representations of deep neural networks are used to cluster scenarios. Also a lot of non-machine learning approaches were used to determine representations or similarities of scenarios, such as in \cite{Wang2020a} (DTW), \cite{Hauer2020a} (DTW+PCA), \cite{Langner2019a} (Dynamic-Length-Segmentation), \cite{Kerber2020a} (custom similarity measure). In this work, an expert-knowledge aided latent representation is introduced, and hence differs from the aforementioned works. To the best of the authors knowledge, this is the first work utilizing a triplet-based autoencoder scheme for the novelty detection of traffic scenarios.

Another viewpoint to compare this work to others is the information used from the scenario. In this work, only the static information is utilized. The most infrastructural information is considered in \cite{Langner2019a}, where categorical and continuos variables are used to roughly describe the environment. In the works \cite{Kruber2018a}, \cite{Kruber2019b} only little static information are used. Whereas in the works \cite{Wang2020a}, \cite{Harmening2020a}, \cite{Demetriou2020a}, \cite{Hauer2020a} and \cite{Kerber2020a} only spatial and dynamic information is used. This work focuses on the static environment of a traffic scenario. Contrary to the former works images of the infrastructure are used here.

In this work, novelty detection is used instead of clustering. While some works focus on the detection of anomalies or corner cases in videos or images (e.\, g. \cite{Hasan2016a},\cite{Bolte2019a}), there are only a few works addressing the detection of novel traffic scenarios through outlier detection on a scenario level. In \cite{Langner2018a} the novelty of traffic scenarios is estimated through an autoencoder. The autoencoder is trained on a data set containing known scenarios. If a scenario is passed through the network, it is assumed to produce a high reconstruction error if it is novel. This reconstruction assumption is a widely used mechanism when performing outlier detection with deep learning.
In this work, it is refrained from using the reconstruction paradigm, rather to learn latent representations which can be used with simpler outlier detection mechanisms. In \cite{Wurst2020a} the novel outlier method ULEF based on directed $k$-nearest neighbor graphs is presented. The method is applied to road infrastructure images. In this work, the road infrastructure image generation as in \cite{Wurst2020a} is used. Moreover, the performance of ULEF will be compared to this work.

\subsection{Triplet Learning and Outlier Detection}
This work is using triplet learning to form the latent space. Triplet networks \cite{Schroff2015a} are used to perform deep metric learning through ranking loss. In tile2vec \cite{Jean2019a}, triplet networks are used for geo-spatial analysis. The sampling is determined based on the distances of the tiles. The architecture used in this work consists of an autoencoder structure. In \cite{Yang2019a}, an autoencoder network is combined with triplet learning. The difference of this work to \cite{Yang2019a} lies in the application, the specific network architecture and the triplet mining.

To aid the detection of outliers, some works are using metric learning. An example is \cite{masana2018metric}, where out-of-distribution data is used for the training.

In the field of deep learning various approaches to detect outliers exist. The most common approach is to use the reconstruction paradigm. For example in \cite{Schlegl2019a}, a GAN is used as generator network, where an additional encoder is trained to learn the mapping from an input image to the GAN's latent representation. An extension of the reconstruction paradigm is using the hidden reconstructions additionally \cite{Kim2020a}.

In this paper, standard outlier detection methods are applied in the latent space, namely the \textit{Local Outlier Factor} (LOF) \cite{Breunig2000a}, \textit{Angle-Based Outlier Detection} (ABOD) \cite{Kriegel2008a}, \textit{Isolation Forest} (IF) \cite{Liu2008a} and the \textit{One-Class Support Vector Machine} (OCSVM) \cite{Schoelkopf2000a}.

\section{METHOD}\label{sec:method}
\begin{figure*}[t]
	\vspace{2mm}
	\centering
	\input{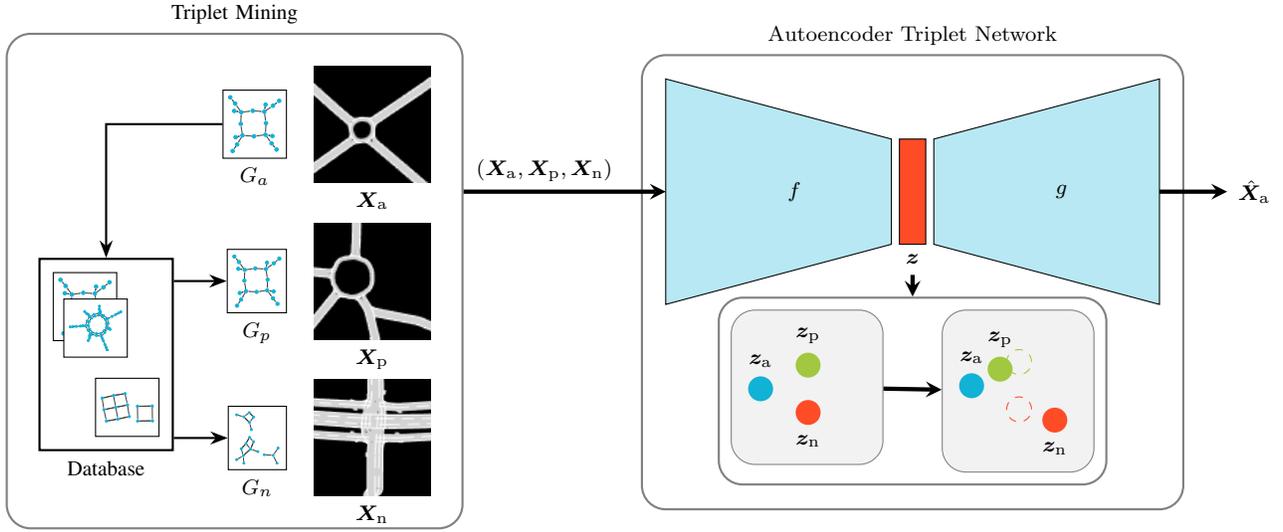}
	\caption{Triplet Mining and Autoencoder Network}
	\label{fig:pipeline}
	\vspace{-0.5cm}
\end{figure*}
In this work, an autoencoder network utilizing triplet loss is used to project road infrastructure images into the latent space, where the outlier detection is performed. By using the proposed architecture, a latent representation for road infrastructure images is learned. During training, the infrastructure is assessed in two ways. First, the visual appearance of the images, hence the shape of roads etc., is considered via the reconstruction loss. Second, the similarity of infrastructure topologies is taken into account via the similarity of their connectivity graphs, allowing data triplets to be identified. A data triplet consists of three data points: the anchor, a sample similar to the anchor and a sample dissimilar to the anchor, which are required for the triplet loss.

This section is split into the following parts. First, the data generation and used similarity measure is explained. Second, the realized triplet autoencoder network as well as the used triplet mining is summarized. Last, the traffic scenario detection in the latent space is described.

\subsection{Data Generation and Similarity Measure}
In this paper, the static part of a traffic scenario is described through its road infrastructure. For this purpose, a black and white image of the infrastructure in top-view is generated. Furthermore, a connectivity graph is generated, which will be required to generate the similarity between infrastructure images.

The used tool-chain consists of the following steps: a) position selection, b) map data collection, c) image generation and d) connectivity graph generation. \textit{OpenStreetMap} (OSM) \cite{OpenStreetMap} is used for the map data. Also the position selection is realized via OSM.

\paragraph{Position Selection}
Within a search area, all OSM nodes can be considered as positions. For this work the valid positions are restricted to public and car-drivable roads. Identifying the nodes and positions can be realized through the API of OSM for example. Each selected position leads to a single data set entry, consisting of an image and the corresponding connectivity graph.

\paragraph{Map Data Collection}
For each selected position, it is required to get the associated map data. In this work, the OSM map with a parameterizable bounding box around the location is converted into an openDRIVE \cite{Dupuis2015a} map, such that the image generation tool as introduced in \cite{Wurst2020a} can be used. For the conversion from OSM to openDRIVE, the netconvert tool of SUMO \cite{SUMO2018} is used.

\paragraph{Image Generation}
The image generation is realized as proposed in \cite{Wurst2020a}. Hence, the roads are colored gray, lane markings white and the background black. Furthermore, all non-public and non-car accessible roads are not rendered. For each position an image is generated.

\paragraph{Graph Generation}
The graph generation is new with respect to \cite{Wurst2020a} and is realized as follows. First, the complete openDRIVE map is converted into a network graph, using the connectivity and neighbor information. The result is somewhat comparable to the routing graph realized in Lanelet2 \cite{poggenhans2018lanelet2}. In the next step, the selected location is assigned to one of the graph's node. Hence, the position on the road is identified. Then, the graph is cropped and simplified using the following rules:
\begin{itemize}
	\item Use all nodes which can be reached within the time $t_\mathrm{max}$, given the allowed speed, up to and including all nodes of the first junction (e.\,g. crossing, roundabout).
	\item Use all nodes which are neighboring lanes.
\end{itemize}

The above steps are used to generate the data set $\mathcal{D}=\left\lbrace(\bm{X}_1,G_1),\dots,(\bm{X}_M,G_M)\right\rbrace$, where $M$ is the number of selected positions, and hence the number of resulting images and graphs. The image for the $m$-th position is given by the matrix $\bm{X}_m \in \mathbb{R}^{S \times S}$ with $S$ the selected resolution of the image. Analogous, the graph for the $m$-th position is given by $G_m=(V_m,E_m)$, where $V_m$ are the vertices and $E_m$ the edges of the graph. The actual selected position is not used after the corresponding image and graph are extracted.

\paragraph{Similarity Measure}
For the triplet-based learning, a notation of being similar or dissimilar is required. In this work, the graphs are used for this purpose. Two cropped areas are considered to be similar if their connectivity graphs are similar. The graphs need to be permuted versions of each other to be similar. This is the case if there exists an isomorphism between the two graphs. Given the graphs $G_i$ and $G_j$, they are similar if there exists a bijection $p:V_i \rightarrow V_j$ such that $(u,v)\in E_i \iff (p(u),p(v))\in E_j$. Using the notation $G_i \cong G_j$ for two graphs being isomorph, the similarity function is defined as
\begin{equation}
	s\left(G_i,G_j\right)=\left\lbrace\begin{array}{ll}
	1 & \text{if }G_i \cong G_j\\
	0 & \text{else}\\
	\end{array}\right. .\label{eq:graphSim}
\end{equation}
This similarity measure considers all data points as same, when their connectivity is the same. Within the triplet mining block of Fig. \ref{fig:pipeline} examples of extracted graphs alongside their images are shown.

\subsection{Triplet Autoencoder}
The main part of this work is the triplet-based autoencoder network. It is used to produce latent representations for given input images of road infrastructures. This section will address the architectural details, the used loss, the triplet mining and details on the used networks.

Triplet learning realizes metric learning via a ranking loss. The objective is to enforce similarity in the latent space based on data triplets. As explained before, each triplet consists of an anchor, a positive example and a negative example. The anchor and the positive example are similar, according to the used similarity measure. Consequently, the negative example is dissimilar to the anchor. For example, in \cite{Schroff2015a}, the anchor is an image of a face, the positive example is another image of the same person and the negative is an image of another person. The objective of the training is to push the latent representation of the negative example away from the latent representation of the anchor while pulling the latent representation of the positive example closer. This way, the metric learning is realized.

In this work, a triplet learning scheme is used, where the road infrastructure images are used as input, with the anchor $\bm{X}_\mathrm{a}$, the positive example $\bm{X}_\mathrm{p}$ and the negative example $\bm{X}_\mathrm{n}$. Let $f$ be a trainable network, realizing the mapping from the input representation to the latent representation $f: \bm{X} \mapsto \bm{z}$ with $\bm{z}\in\mathbb{R}^L$ being the latent representation with dimensionality $L$. During the training, each sample of the triplet $(\bm{X}_\mathrm{a},\bm{X}_\mathrm{p},\bm{X}_\mathrm{n})$ is passed through $f$ separately, leading to the latent triplet $(\bm{z}_\mathrm{a},\bm{z}_\mathrm{p},\bm{z}_\mathrm{n})$. During inference, only single samples will be passed through $f$. The training of $f$ is partially realized by using the triplet loss
\begin{equation}\label{eq:tripletLoss}
	\mathcal{L}_\mathrm{tri}(\bm{X}_\mathrm{a},\bm{X}_\mathrm{p},\bm{X}_\mathrm{n}) = \max\left(\alpha + d_\mathrm{ap} - d_\mathrm{an},0\right),
\end{equation}
with the squared distance between the anchor representation and the positive example representation $d_\mathrm{ap}=\vert\vert f(\bm{X}_\mathrm{a}),f(\bm{X}_\mathrm{p})\vert\vert_2^2$, the squared distance between the anchor representation and the negative example representation $d_\mathrm{an}=\vert\vert f(\bm{X}_\mathrm{a}),f(\bm{X}_\mathrm{n})\vert\vert_2^2$ and the margin $\alpha$. The triplet loss' objective is twofold, to lower the distance between the positive example and the anchor $d_\mathrm{ap}$ while simultaneously increase the distance between the negative example and the anchor $d_\mathrm{an}$ until $d_\mathrm{ap}+\alpha$. This way the latent representation is forced to follow the definition of similarity as provided by the triplets.

One key point of the triplet scheme is the triplet mining. While sampling $\bm{X}_\mathrm{a}$ is realized by randomly picking an image from the data set $\mathcal{D}$, determining the positive and negative examples is based on the anchor. The similarity of two road infrastructure images is defined through Eq. \ref{eq:graphSim}. Hence, the positive example is randomly picked out of all samples having the same connectivity as the anchor. Given the graph of the anchor as $G_\mathrm{a}$ the graph of the positive example $G_\mathrm{p}$ is picked from $\mathcal{G}_\mathrm{p}=\left\lbrace G \vert s(G,G_\mathrm{a})=1\right\rbrace$, given $ G \in \mathcal{G}$, where $\mathcal{G}$ is the set of all graphs of $\mathcal{D}$. Using the same notation, the graph of the negative example $G_\mathrm{n}$ is picked from $\mathcal{G}_\mathrm{n}=\left\lbrace G \vert s(G,G_\mathrm{a})=0\right\rbrace$.

The selection of especially the negative sample has a significant influence on the training. Considering the case where the negative is already too far away and hence there is no training contribution for this triplet, since $d_\mathrm{an} \geq d_\mathrm{ap}+\alpha$ would lead to $\mathcal{L}_\mathrm{tri} = 0$. Such samples are called easy negatives. On the other hand-side having hard negatives, i.\,e. $d_\mathrm{an} < d_\mathrm{ap}$, might lead to bad local minima early in the training \cite{Schroff2015a}. Both types should be prevented. Therefore, the objective is to sample data points that are called semi-hard negative samples. They are further away than the positive example but still within the margin $\alpha$, such that $d_\mathrm{ap} < d_\mathrm{an} < d_\mathrm{ap}+\alpha$ holds.

The triplet loss conditioned training might be sufficient to separate different connectivity types, for example roundabout with 4 versus 3 exit roads. However, another objective in this work is to ensure that neighboring points in the latent space have visually very similar input images.
The reasoning for this is twofold. First, it is assumed that the shape of the road also has an influence on the novelty of a scenario. Furthermore, it is assumed, the more similar the neighborhood in the latent space is, the more reliable the projection is for unknown data. By this, even within the same connectivity group further refinement is achieved. For this purpose, a decoder is introduced into the overall architecture. The decoder is used to regularize the latent space, such that it can be used to reconstruct the anchor $\bm{X}_\mathrm{a}$. The underlying assumptions is, that for the reconstruction to work, the latent space has to be formed in such away, that the neighbors are not sharing only connectivity-based similarities but also visual similarities.
The trainable decoder network $g$ is used to generate the reconstructed anchor image $\hat{\bm{X}}_\mathrm{a}$ by $g:\bm{z}\mapsto\hat{\bm{X}}$. The reconstruction loss of the anchor,
\begin{equation}
	\mathcal{L}_\mathrm{rec}(\bm{X}_\mathrm{a}) = \vert\vert \bm{X}_\mathrm{a} - g\left(f\left(\bm{X}_\mathrm{a}\right)\right)\vert\vert_2^2
\end{equation}
is used to enforce the above described objective.

The complete architecture is trained using the loss
\begin{equation}
	\mathcal{L} = \mathcal{L}_\mathrm{tri}+\mathcal{L}_\mathrm{rec},
\end{equation}
such that both objectives are fulfilled: connectivity-based structure refined by visual similarity. The overall pipeline of the triplet autoencoder network including the triplet mining can be seen in Fig. \ref{fig:pipeline}. The exemplary graphs are the results for the shown images. This highlights the need for the local refinement by the decoder, since the shown anchor and its positive example are sharing the same connectivity but have quite different visual appearance. Below the network, the learning process indicated by the triplet loss is symbolized.

The triplet autoencoder network can be trained on a huge data set such as OSM, ensuring a strong projection method. After training, the network can be used during inference phase to project new data. This way, it is possible to train the network only once. Furthermore, it allows one to store the data in a compressed format, i.\,e. the latent representations.

The decoder network is kept fairly simple in order to limit the complexity of the latent space. The decoder network structure is kept the same for all experiments, only the encoder is varied. For the encoder network the ResNet and ViT are tested and are therefore briefly explained in the following.

\paragraph{ResNet}
In \cite{He2016a} the widely used ResNets 18, 50 etc. are proposed. Residual networks consist of multiple residual blocks with multiple convolutional layers each. The input to each residual block is added to the output again. Since those networks are widely known, further details will not be given here but can be found in \cite{He2016a}.

\paragraph{Vision Transformer}
Most recently, ViTs have been introduced in \cite{Dosovitskiy2021a}, using the attention mechanism-based transformers as in \cite{Vaswani2017a}. ViT is a convolution-free network for image classification, showing state-of-the-art performance.

\begin{figure}
	\vspace{2mm}
	\centering
	\input{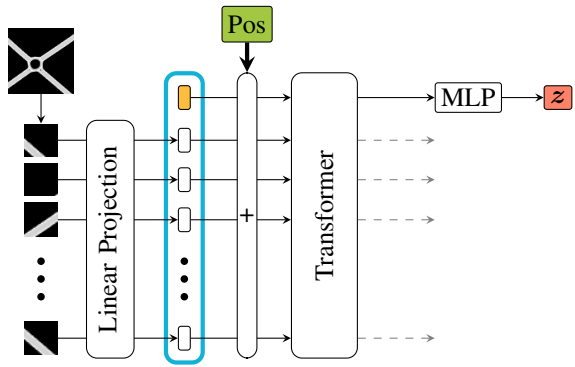}
	\caption{Vision Transformer as used in this work. Inspired from \cite{Dosovitskiy2021a}.}
	\label{fig:ViT}
	\vspace{-0.5cm}
\end{figure}

The transformer usually uses an encoder and a decoder for sequence to sequence modelling, but the ViT is using only the encoder part, as shown in Fig. \ref{fig:ViT}. First, an image is split into flattened patches which are linearly projected, leading to the input embeddings (marked blue). Then, an additional embedding token (orange) is concatenated to the input embeddings before added to the learnable positional embedding (green). The sum is fed as an input to the transformer. Inside the transformer, a multi-layer multi-head-attention network is realized.

The output token, corresponding to the additional embedding token, is fed through a \textit{Multi Layer Perceptron} (MLP). Here, in difference to the original ViT, the output vector is the latent representation $\bm{z}$ (red), whereas in ViT it is used to predict a class label.

\subsection{Novelty Detection}
The overall objective in this work is to detect unknown road infrastructures. For this purpose, outlier detection methods can be applied. Let the base data set $\mathcal{D}_\mathrm{base}$ be the data already known, for example, the scenarios an autonomous driving function has already been tested on. If a data point is tested with respect to its outliership it is basically investigated if the point fits into the known data $\mathcal{D}_\mathrm{base}$. The training data $\mathcal{D}$, used to train the projection $f$, is not required to be the same as the base data $\mathcal{D}_\mathrm{base}$.

Provided with the latent infrastructure representation $\bm{z}$, simple outlier mechanisms can be applied directly on the latent space instead of the input space. This way, the enforced similarity utilizing connectivity and shape will mainly be responsible for detecting outliers. In the following, let $o$ be the outlier ranking function fitted on $\mathcal{D}_\mathrm{base}$, mapping each input to an estimated outlier score $v=o(\bm{z})$.

Within this work various outlier detection methods are used. Methodological details are not covered in this work. Interested readers may refer to the corresponding works. In the following the used methods are briefly summarized.
\paragraph{Local Outlier Factor}
The LOF \cite{Breunig2000a} is analyzing how isolated a data point is with respect to its neighbors. For this the local densities are used.

\paragraph{Isolation Forest}
The IF \cite{Liu2008a} is estimating the outlierness through the number of splits required to isolate a data point, given randomly grown trees. It holds, that the fewer splits the more outlying.

\paragraph{Angle-Based Outlier Detection}
Another neighborhood-based approach is ABOD \cite{Kriegel2008a}. The variance of the direction vector's angle from the point under investigation to all other data points is investigated. The higher the variance the lower the outlierness. In this work, the fast version of ABOD is used, considering only the $k$ nearest neighbors of the investigated point.

\paragraph{One-Class Support Vector Machine}
The OCSVM \cite{Schoelkopf2000a} is the extension of the normal SVM to the problem of outlier detection. The data is mapped to the feature space as by the kernel. In the feature space, the data is separated from the origin via a hyperplane. This way, the decision boundary is meant to encapsulate the data in the input space.

\paragraph{UMAP-based Local Entropy Factor}
In \cite{Wurst2020a} the neighborhood of data point's is used to define local similarity measures. A points outliership is evaluated based on how well it fits into its neighbors' neighborhoods using the entropy and the point-wise similarity.

\FloatBarrier
\section{EXPERIMENTS}\label{sec:exps}
The introduced method is evaluated in this section. Various architecture realizations and outlier detection methods are compared. This section is split into the following parts. First, the details about the data used for the outlier detection and the data used for visual analysis is explained. Second, the analyzed architectures are briefly summarized. The outlier detection performance is discussed in the third part. The various resulting latent space visualizations are shown and discussed in the fourth part. In the fifth subsection, the local visual similarity quality is assessed, proposing another possibility to evaluate the performance of the shown architectures. The results are summarized in the last part.

\subsection{Data sets}
The training data set $\mathcal{D}$ is decoupled from the base data set used for the outlier detection $\mathcal{D}_\mathrm{base}$. Both are briefly described in the following. For both the images are showing a region of size $\unit[100]{m}\times\unit[100]{m}$, while the reachable time for generating the graphs is selected as $t_\mathrm{max}=\unit[5]{s}$.

The training data set $\mathcal{D}$ consists of $\approx 70\,000$ pairs of images and graphs. To provide an insight to the data set, it has been analyzed with respect to rough groups, but those groups are not used in the training. In total, approx. 13\,400 highway, 16\,900 roundabouts, 18\,000 crossings, 19\,900 single lane and 1\,700 multiple lane non-highway pairs are used. The extraction region for the highway and roundabout pairs is selected to be the complete district of Upper Bavaria in Germany. The extraction region for the remaining types is the city of Ingolstadt in Germany with its adjacent counties.

The base data set $\mathcal{D}_\mathrm{base}$ to fit the models of the outlier detection methods is taken from \cite{Wurst2020a}.
Therefore, the data considered to be known are highway images only. The outlying test data is taken from \cite{Wurst2020a} as well. It consists of rural images and inner-city images, being gathered in the anomaly data set $\mathcal{D}_\mathrm{ano}$. It holds that $\mathcal{D}_\mathrm{base} \in \mathcal{D}$ and $\mathcal{D}_\mathrm{ano} \in \mathcal{D}$.

\subsection{Architectures}
\begin{table*}
	\vspace{2mm}
	\caption{Outlier Detection Performance -- AUC}\label{tb:ODP}
	\scriptsize
	\centering
	\begin{tabular}{c|c|c|c||c|c|c|c|c||c|c|c}
		\multicolumn{3}{c|}{Architecture}&\multirow{2}{*}{OD Input}&\multirow{2}{*}{LOF}&\multirow{2}{*}{ULEF}&\multirow{2}{*}{IF}&\multirow{2}{*}{OCSVM}&\multirow{2}{*}{ABOD}&\multirow{2}{*}{CAE-ORD}&\multirow{2}{*}{CAE-SAP}&\multirow{2}{*}{f-AnoGAN}\\
		\cline{1-3}
		$f$	& $g$			& $\mathcal{L}_\mathrm{tri}$			& & & &  & & & & & \\
		\hline
		\hline
		Res-S	&\xmark		& \checkmark	& $\bm{z}$ & $0.913$ 			& $0.775$ & $0.935$ & $\mathbf{0.959}$ 	& $0.892$ 	& -- & -- & --\\
		\hline
		Res-S	&\checkmark	& \checkmark	& $\bm{z}$ & $\mathbf{0.954}$ 	& $0.696$ & $0.933$ & $0.950$ 			&$\mathbf{0.954}$		& -- & -- & --\\
		\hline
		Res-18	&\checkmark	& \xmark		& $\bm{z}$ & $0.696$ 			& $0.776$ & $0.770$ & $0.752$ 			& $\mathbf{0.784}$	& -- & -- & --\\
		\hline
		Res-18	&\checkmark	& \checkmark	& $\bm{z}$ & $0.949$ 			& $0.781$ & $0.917$ & $0.912$ 			&$\mathbf{0.956}$		& -- & -- & --\\
		\hline
		ViT-S	&\checkmark	& \checkmark	& $\bm{z}$ & $0.730$ 			& $0.689$ & $0.698$ & $0.910$ 			& $\mathbf{0.920}$	& -- & -- & --\\
		\hline
		ViT-L	&\checkmark	& \checkmark	& $\bm{z}$ & $0.900$ 			& $0.793$ & $0.707$ & $0.937$ 			& $\mathbf{0.956}$ 	& -- & -- & --\\
		\hline
		\hline
		\multicolumn{3}{c|}{--}				& $\bm{X}$ & $0.446$ 			& $0.612$ & $0.196$ & $0.247$ 			& $0.700$ 			& $0.855$ & $0.845$ & $0.758$\\
		\hline
	\end{tabular}
	\vspace{-0.5cm}
\end{table*}
In order to investigate the influence of various architecture realizations, different versions of the network were used for the experiments. The architecture of the decoder is kept the same for all the experiments, enabling a fair comparison. The encoder is realized through different networks. Furthermore, the loss function was changed for some experiments, highlighting the importance of the overall pipeline. Here, the various used options will be briefly summarized.

For all architectures, the following parameters hold: image size $64 \times 64$, epochs $200$, latent dimensionality $50$.
\paragraph{Res-S}
A small ResNet, as ResNet-18 but consisting of fewer parameters. Learnable parameters $\approx 0.3\cdot10^6$.

\paragraph{Res-18}
A larger ResNet. Here the basic implementation of ResNet-18 \cite{He2016a} is adjusted for single channel inputs. Learnable parameters $\approx 11.0\cdot10^6$.

\paragraph{ViT-S}
A small version of the ViT. Learnable parameters $\approx 0.2\cdot10^6$. (patch size $8\times8$, layers $6$, dim. of input embedding $n_\mathrm{patches}\times64$, internal MLP dim. $128$).

\paragraph{ViT-S}
A large version of the ViT. Learnable parameters $\approx 6.7\cdot10^6$. (patch size $8\times8$, layers $20$, dim. of input embedding $n_\mathrm{patches}\times256$, internal MLP dim. $128$).

Furthermore, \xmark $g$ is highlighting architectures where the decoder network is deactivated for the experiments. Therefore, the loss is only based on the triplet part. The triplet loss is not used for the architectures marked with \xmark $\mathcal{L}_\mathrm{tri}$. Hence, only the reconstruction loss is used.

\subsection{Novelty Detection}
The various architectures are evaluated with respect to their novelty detection performance. For this, the outlier detection methods LOF, ULEF, OCSVM and ABOD are applied in the latent space of the resulting network. The outlier detection methods are also applied in the input space to highlight the performance gain when using the latent representation (last row in Tb. \ref{tb:ODP}). The networks are trained using $\mathcal{D}$, while the outlier detection is performed using $\mathcal{D}_\mathrm{base}$ as known and considering the remaining $\mathcal{D}_\mathrm{ano}$ as unknown. Therefore, the novelty detection is applied to the example where only highway images $\mathcal{D}_\mathrm{base}$ are known. Then unknown data points, such as inner-city images, $\mathcal{D}_\mathrm{ano}$ are investigated with respect to their outliership. If the outlier detection is able to identify that for example inner-city images are outlying with respect to highway images, the novelty detection task is fulfilled successfully. This evaluation shows the outlier detection capabilities with respect to connectivity classes.

Additionally, the reconstruction-based outlier detection methods f-AnoGAN \cite{Schlegl2019a} and RaPP \cite{Kim2020a} are evaluated. For the RaPP, a convolutional autoencoder is trained using $\mathcal{D}_\mathrm{base}$ then the normal reconstruction error is used for outlier detection (CAE-ORD) and the simple aggregation along pathway (CAE-SAP) as introduced in \cite{Kim2020a}. For the f-AnoGAN the network is also trained only on $\mathcal{D}_\mathrm{base}$.

In Tb. \ref{tb:ODP}, the resulting \textit{Area Under Curve} (AUC) values are shown for the various combinations. The AUC will be 1 if all outliers are detected correctly. As a result of the experiments, it is shown that when the latent representations are used for the outlier detection, the performance of each method is improving (LOF, ULEF, IF, OCSVM, ABOD). The networks are providing a powerful latent representation, where simple outlier detection methods can perform well. In fact, most combinations (network with a basic outlier detection method) are outperforming other baseline methods such as CAE-SAP and f-AnoGAN. Using one of the architectures in combination with ABOD is the preferable solution, since it yields the highest performance for all triplet autoencoding-based schemes (\checkmark$g$ and \checkmark$\mathcal{L}_\mathrm{tri}$) and provides the highest performance overall. The results show that the proposed approach to include domain-knowledge in shaping the latent space improves the outlier detection performance significantly in this application.

The relevance of the triplet loss becomes clear when comparing the simple autoencoder architecture (Res-18 \xmark $\mathcal{L}_\mathrm{tri}$) against the one including the triplet loss (Res-18 \checkmark $\mathcal{L}_\mathrm{tri}$). The use of the triplet loss increases the performance for all architectures remarkably. The influence of the decoder can be identified from Res-S \xmark$g$ versus \checkmark$g$. Its contribution to the outlier detection is not as clear as for the triplet loss. Indeed, for some methods the outlier detection is getting slightly worse, however, for some it is getting better. The reason for introducing the decoder is the local visual similarity, which is not represented by the outlier detection analysis, since this is only covering the class oriented scale. For this purpose, another analysis will be carried out in the following section.
Further comparison can be drawn from the different encoder types when using the triplet loss and a decoder.
Because of its stable and high performance only ABOD is considered for further discussions. The Res-18 is only slightly better than the smaller Res-S.
The performance difference for the two ViT version is more significant. Here, the ViT-L would be the architecture of choice. In conclusion, using the triplet loss as well as the decoder is clearly beneficial, while either the Res-S, Res-18 or ViT-L can be used from the outlier detection performance point of view.
\subsection{Latent Space Visualization}
\begin{figure}[h]
	\vspace{2mm}
	\centering
	\setlength\tabcolsep{1pt}
	\begin{tabular}{c c c}
		\includegraphics[width=0.3\columnwidth]{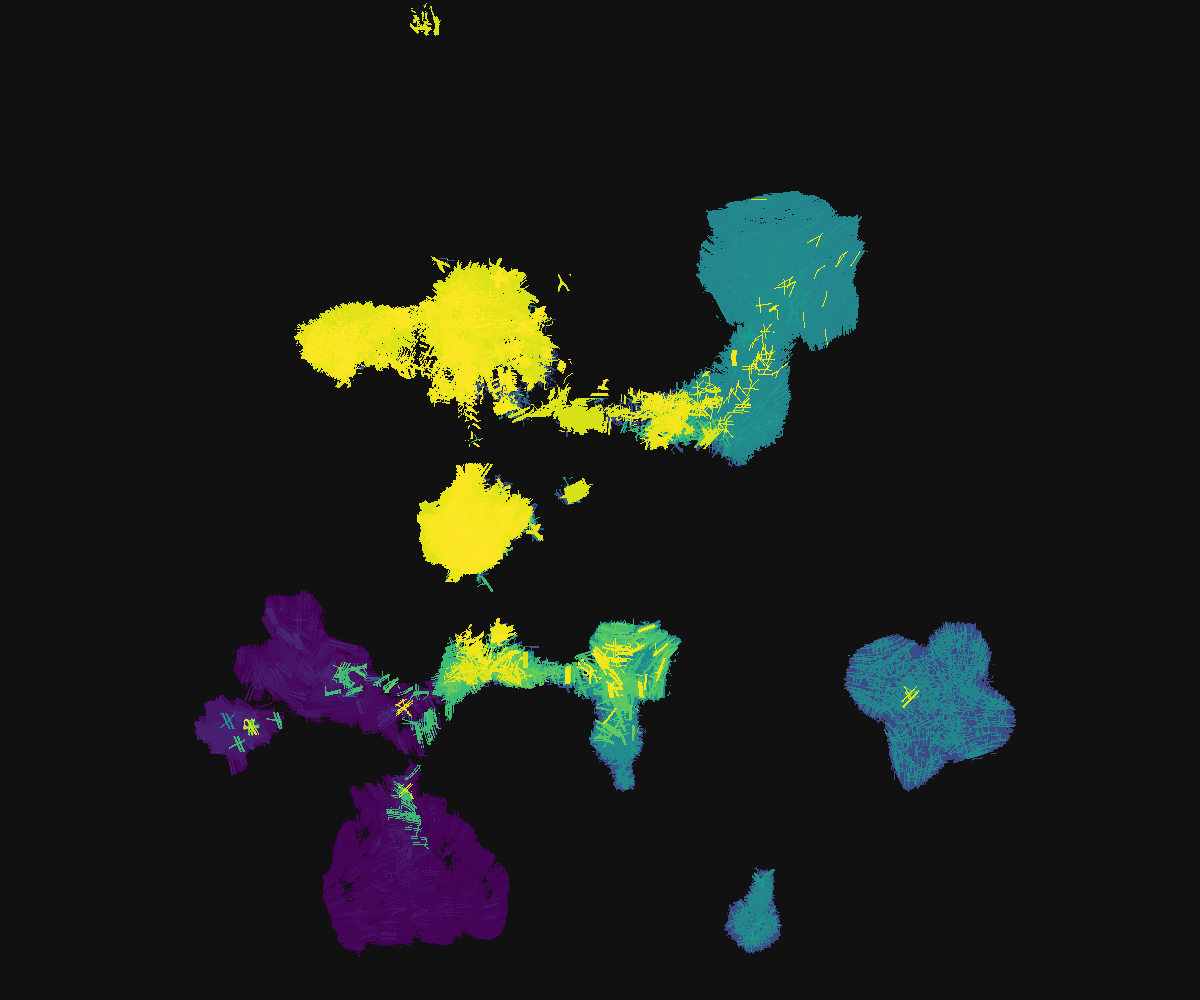} & \includegraphics[width=0.3\columnwidth]{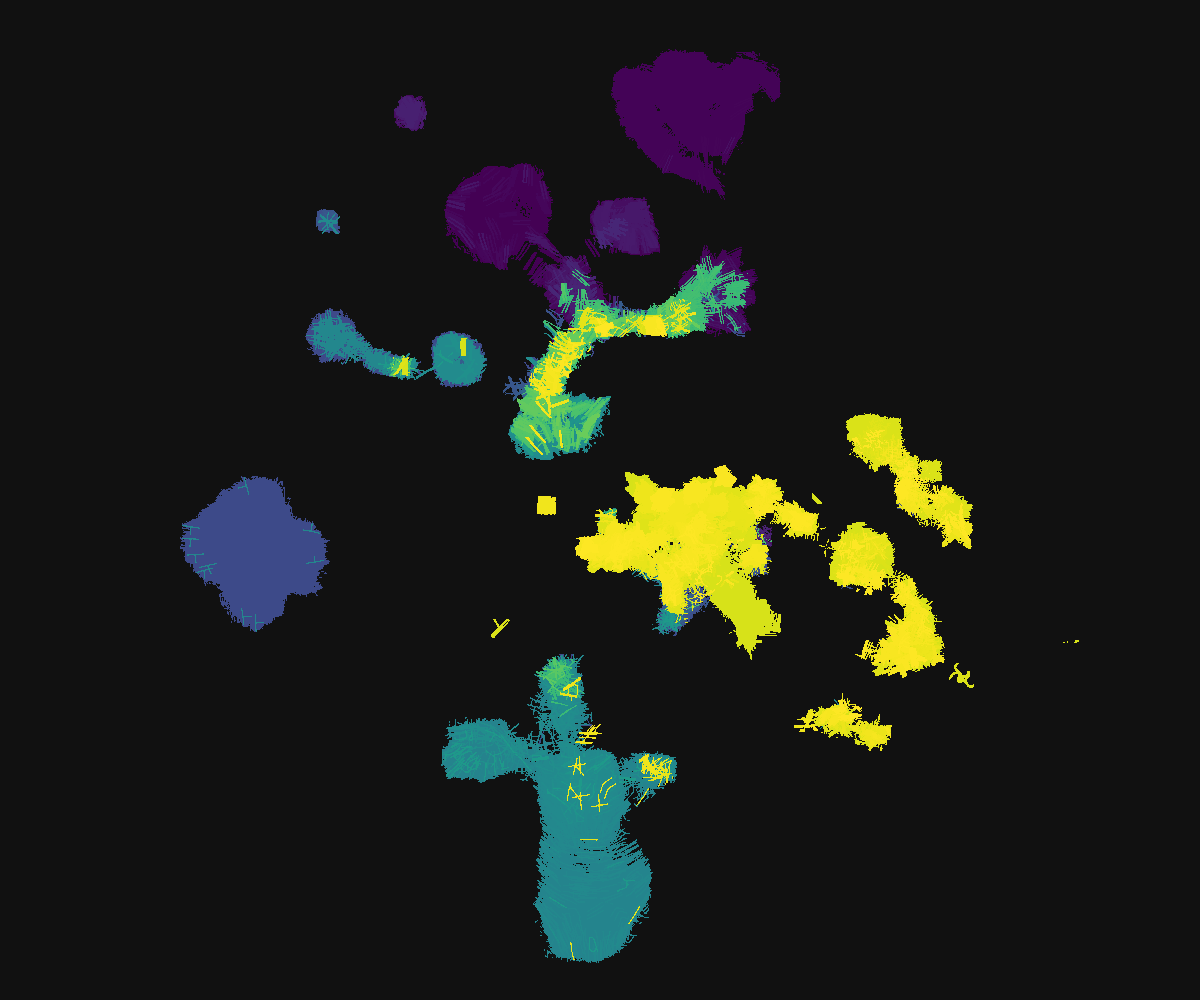} & \includegraphics[width=0.3\columnwidth]{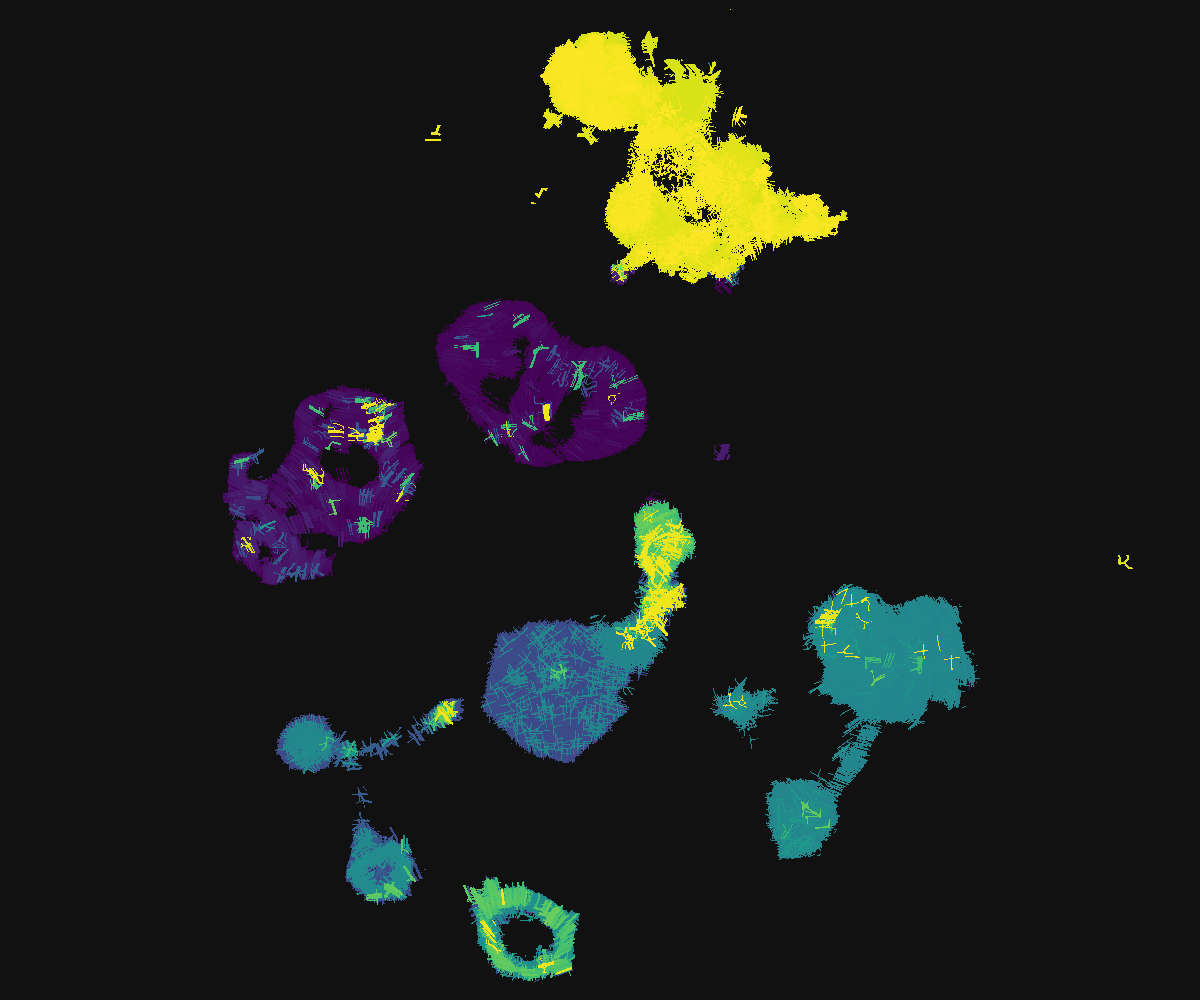} \\
		Res-S \checkmark $g$ & Res-18 \checkmark $\mathcal{L}_\mathrm{tri}$ & ViT-L\\
		\includegraphics[width=0.3\columnwidth]{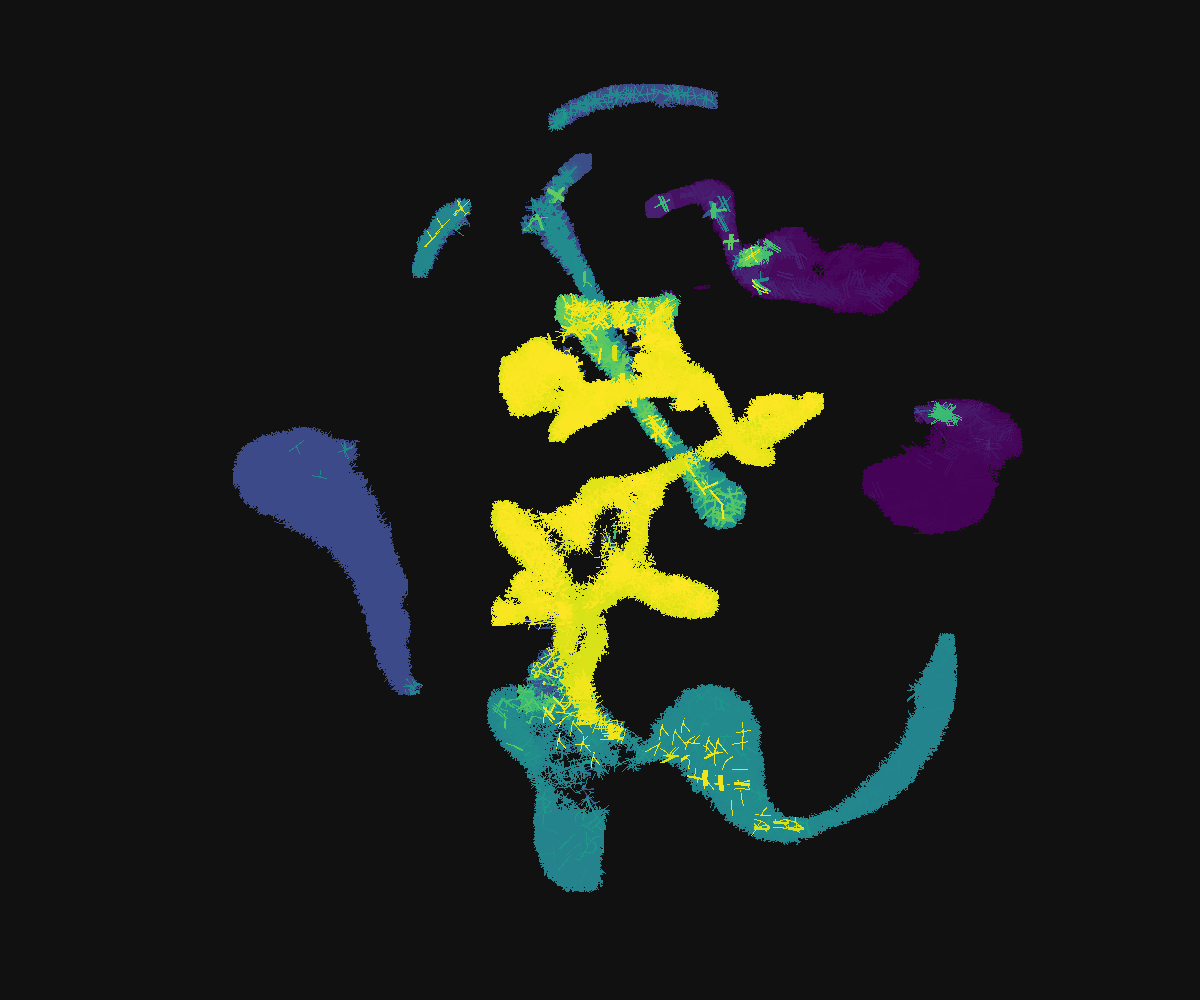} & \includegraphics[width=0.3\columnwidth]{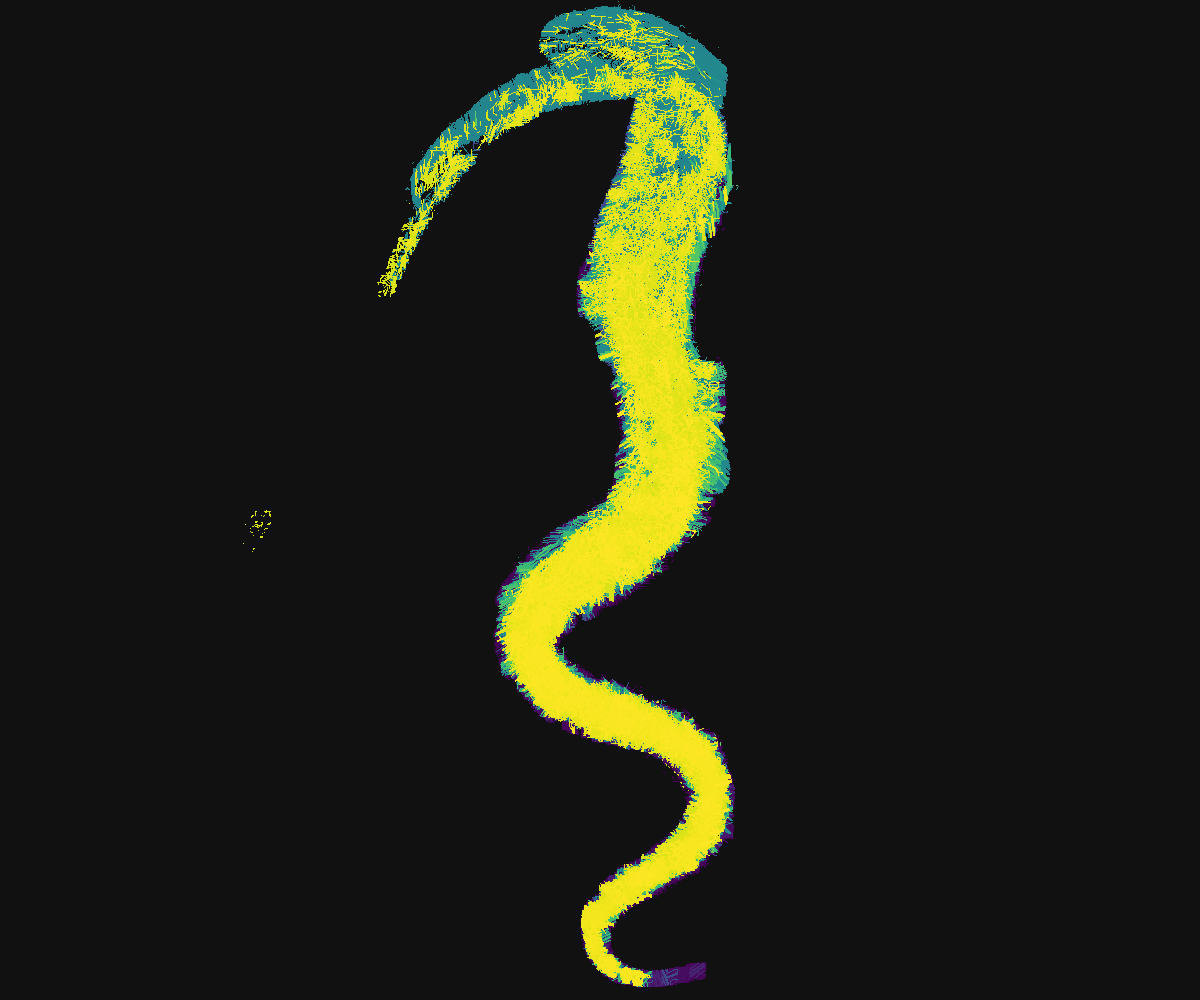} & \includegraphics[width=0.3\columnwidth]{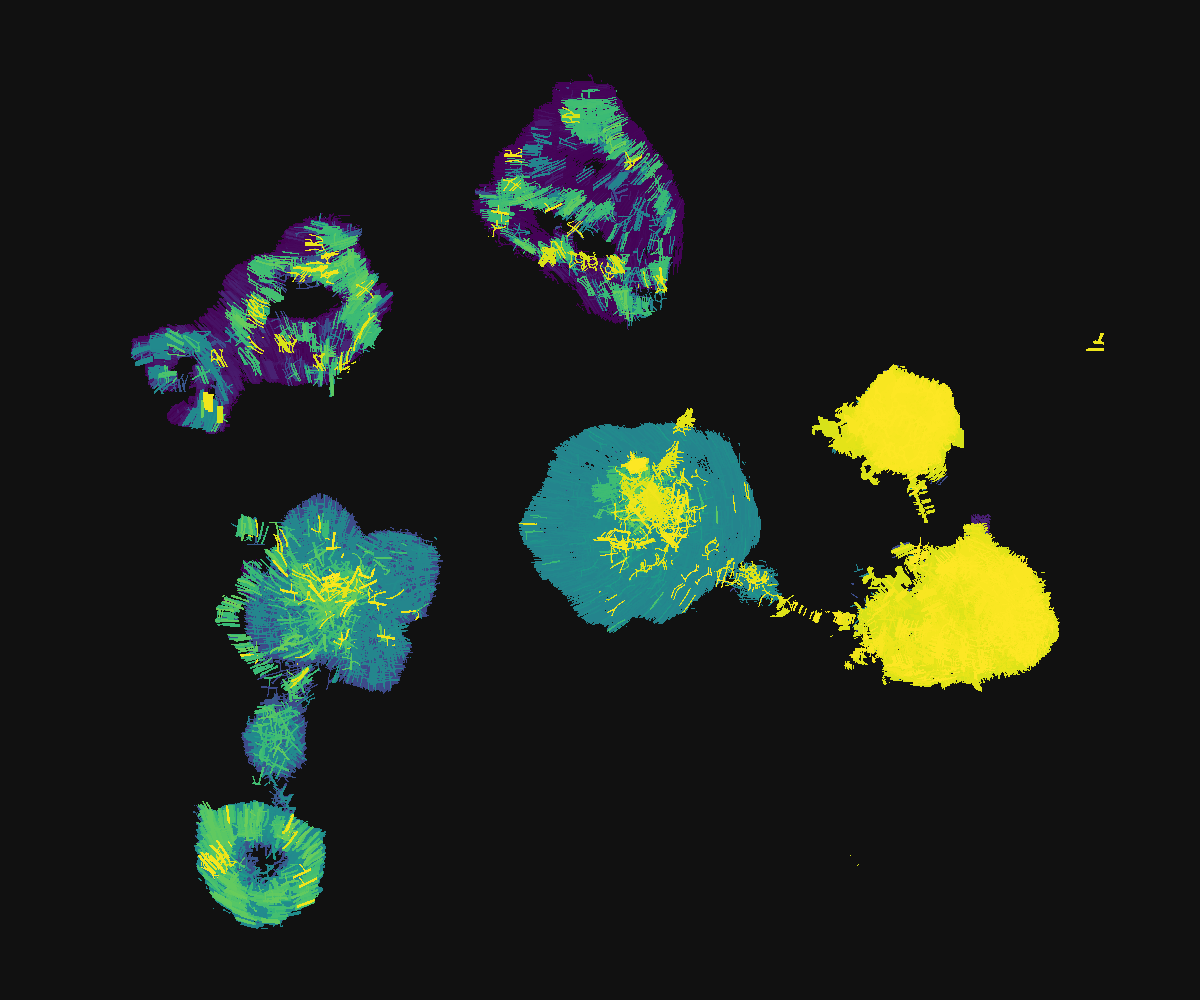}\\
		Res-S \xmark $g$ 	 & Res-18 \xmark $\mathcal{L}_\mathrm{tri}$ 	& ViT-S\\
	\end{tabular}
	\setlength\tabcolsep{6pt}
	\caption{UMAP projections of $\mathcal{Z}$. \resizebox{3ex}{1.5ex}{\includegraphics{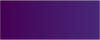}} highway, \resizebox{3ex}{1.5ex}{\includegraphics{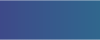}} crossing, \resizebox{3ex}{1.5ex}{\includegraphics{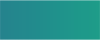}} single lane, \resizebox{3ex}{1.5ex}{\includegraphics{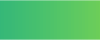}} multiple lane, \resizebox{3ex}{1.5ex}{\includegraphics{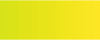}} roundabout.\\
	Interactive visualization tool SCENATLAS: \url{https://jwthi.github.io/SCENATLAS/}}
	\label{fig:vis}
	\vspace{-0.5cm}
\end{figure}
Another approach to access the quality of the latent space is provided through visual investigation. For this purpose, the latent representations $\mathcal{Z}=\left\lbrace\bm{z}_1,\dots,\bm{z}_M\right\rbrace$ are projected into a two-dimensional space using UMAP \cite{McInnes2018a}. In Fig. \ref{fig:vis}, the projections for the various architectures are shown.

On this scale, the difference of Res-S \checkmark $g$ versus \xmark $g$ is hard to figure out. That difference will be further investigated via the local similarity analysis. For the difference produced by the triplet loss (middle column in Fig \ref{fig:vis}), this scale is sufficient. It is visible, that training the autoencoder without the triplet loss leads to a less separable latent representation. The difference between the Res-S and Res-18 (upper) is mainly in the more diffuse distribution of the roundabouts when using Res-18. When comparing the ViT-L against ViT-S, the larger one is showing clear advantage, since in contrary to the smaller version, it is able to distinguish between highway and multiple lane. Comparing the architectures Res-S (up) Res-18 (up) and ViT-L, all of them show clear grouping of the analysis classes, and hence, from this perspective are equally well suited. Interested readers may refer to \url{https://jwthi.github.io/SCENATLAS/}, where the interactive visualization tool SCENATLAS is provided, which allows to discover the latent spaces more intuitively.

\subsection{Local Similarity Analysis}
\begin{figure}[h]
	\vspace{2mm}
	\centering
	\input{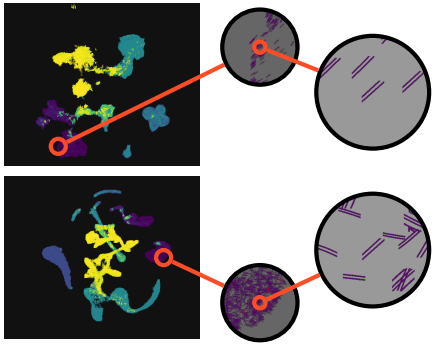}
	\caption{Local visual similarity motivation. Res-S Up: \checkmark g, Down: \xmark g}
	\label{fig:mag}
	\vspace{-0.3cm}
\end{figure}
As stated above, the main motivation of introducing the decoder is the local visual similarity of the latent space. This property has not yet been analyzed. So far, only the more class orientated outlier performance was evaluated, i.\,e. differentiate highway versus non-highway. Therefore, another analysis is performed for the local scale. In Fig. \ref{fig:mag} the problem is visualized. It is showing two zoom levels of the embeddings with (up) and without (down) the decoder. Therefore, the embedding with the decoder appears to be more visually similar. In the following this quantity is assessed through numerical evaluation.

\begin{figure}[t]
	\vspace{2mm}
	\centering
	\begin{tikzpicture}
\footnotesize
\begin{axis}[
    grid=major,
    width=0.9\columnwidth,
    height=0.7\columnwidth,
    xlabel=$k$,
    ylabel=$d_{\mathrm{local}}\downarrow$,
    ymin=0, 
    ymax=30,
    restrict x to domain=0:25,
    xmin=0,
    xmax=25,]

    \addplot[draw=ownBlue,color=ownBlue,mark=none,thick,dashdotted] table [x=x,y=y]{res_18_no_trip.dat};\label{pl:no_trip}
    \addplot[draw=ownBlue,color=ownBlue,mark=none,thick] table [x=x,y=y]{res_18.dat};\label{pl:res_18}
    \addplot[draw=ownGreen,color=ownGreen,mark=none,thick] table [x=x,y=y]{res_small.dat};\label{pl:res_small}
    \addplot[draw=ownGreen,color=ownGreen,mark=none,thick,dashed] table [x=x,y=y]{res_small_no_recon.dat};\label{pl:res_small_no_recon}
    \addplot[draw=ownRed,color=ownRed,mark=none,thick] table [x=x,y=y]{trans_small.dat};\label{pl:trans_small}
    \addplot[draw=ownOrange,color=ownOrange,mark=none,thick] table [x=x,y=y]{trans_big.dat};\label{pl:trans_large}

\end{axis}
\end{tikzpicture}
	\caption{Average distance of the data points to their nearest neighbors corresponding to the respective latent representations. Smaller distance values are better.
	\ref{pl:no_trip}: Res-18 \xmark $\mathcal{L}_\mathrm{tri}$,\ref{pl:res_18}: Res-18, \ref{pl:res_small}: Res-S, \ref{pl:res_small_no_recon}: Res-S \xmark $g$, \ref{pl:trans_small}: ViT-S and \ref{pl:trans_large}: ViT-L}
	\label{fig:roadResults}
	\vspace{-0.5cm}
\end{figure}
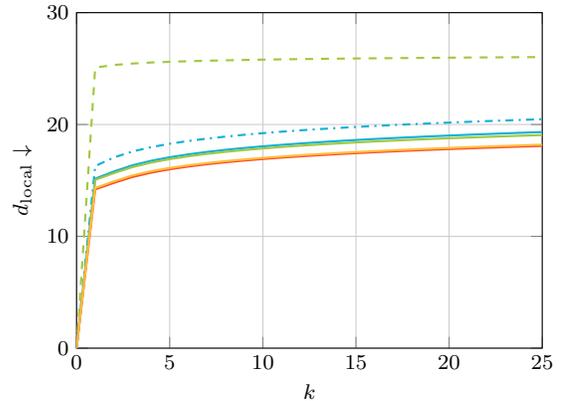

Given the latent representations $\mathcal{Z}$, the $k$ nearest neighbors for the $i$-th sample in the latent space are determined, leading to the nearest neighbor set $\mathcal{K}_i$. Then, the average distance in the input space between a point and its neighbors is determined. By using
\begin{equation}
	d_{\mathrm{local}}(k) = \frac{1}{M}\sum_{m=1}^M\frac{1}{k}\sum_{j\in{\mathcal{K}_i}}\Vert \bm{X}_i - \bm{X}_j\Vert_2
\end{equation}
the average distance between the latent space neighbors is evaluated. Since a low value states a high average visual similarity between the $i$-th point and its neighbors, it is considered that lower values are better. In Fig. \ref{fig:roadResults} the resulting values are shown for all the architecture variants. Therefore, the reasoning for the usage of the decoder is clearly provided. Moreover, the usage of the triplet loss is increasing the performance as well. The Res-S and Res-18 are again on a comparable level, but the ViT based architectures outperform all others. This indicates, that the ViT based method is preferable to identify local, shape based outliers.

\subsection{Summary}
\begin{table}[h]
	\vspace{2mm}
	\centering
	\caption{Architecture Overview}
	\label{tb:archiOverview}
	\begin{tabular}{c|c|c|c|c|c}
		\multicolumn{3}{c|}{Architecture}&\multirow{2}{*}{AUC $\uparrow$}&\multirow{2}{*}{$d_\mathrm{local}(5)$ $\downarrow$}&\multirow{2}{*}{Nparams}\\
		\cline{1-3}
		$f$	& $g$ 	& $\mathcal{L}_\mathrm{tri}$ 		& & & \\
		\hline
		\hline
		Res-S	&\xmark		& \checkmark	&$0.892$& $25.61$ & $0.3\cdot10^6$\\
		\hline
		Res-S	&\checkmark	& \checkmark	&$0.954$& $16.90$ & $0.3\cdot10^6$\\
		\hline
		Res-18	&\checkmark	& \xmark		&$0.784$& $18.28$ & $11.0\cdot10^6$\\
		\hline
		Res-18	&\checkmark	& \checkmark	&$\mathbf{0.956}$& $17.07$ & $11.0\cdot10^6$\\
		\hline
		ViT-S	&\checkmark	& \checkmark	&$0.920$& $\mathbf{16.00}$ & $0.2\cdot10^6$\\
		\hline
		ViT-L	&\checkmark	& \checkmark	&$\mathbf{0.956}$ & $16.12$& $6.7\cdot10^6$\\
		\hline
	\end{tabular}
	\vspace{-0.2cm}
\end{table}
To sum up this analysis, the important values are gathered in Tb. \ref{tb:archiOverview}. Here, the results using ABOD are used. The overall best performance is provided by the ViT-L. However, also the small networks Res-S and ViT-S perform remarkably well. The final suggestion is to use ViT-L with ABOD to achieve high outlier detection performance as well as high visual similarity in the latent space.

\section{CONCLUSION}\label{sec:conc}
A method to identify novel traffic scenarios based on their static component is presented in this work. The introduced pipeline is outperforming existing outlier detection methods. It has the additional advantage that it can be trained on a huge data set (e.\,g. OSM), such that no retraining is necessary. In contrary, methods relying on the reconstruction paradigm would require retraining when detecting unknown scenarios. This work presents a possibility to incorporate expert-knowledge of a scenario's static environment for shaping the latent space of a triplet autoencoder. The presented results show that methods like outlier detection can significantly benefit from a latent space shaped in this way.

Another approach to detect unknown infrastructure based on their topology, could be realized through a simple categorization logic using the graphs, leading to a mixture of experts. However, the presented method provides an insight to the relationship between infrastructure types and additionally the local shape similarity.

Future work might focus on the inclusion of further parts of a scenario.
Given this latent space, a possible improvement in clustering performance can also be investigated.
Another possible extension is to include more infrastructural information into the graphs.

In conclusion, the suggested pipeline consists of a connectivity graph-based similarity definition, an autoencoder triplet network with a ViT as encoder and the ABOD method performing the novelty detection in the latent space. It shows superior performance with respect to its novelty detection capabilities and with respect to the neighborhood similarity evaluation. The interactive visualization SCENATLAS of the latent spaces is provided (\url{https://jwthi.github.io/SCENATLAS/}) and the code implementing the method is made publicly available.

\section{ACKNOWLEDGEMENT}
The authors acknowledge the support by the ZF Friedrichshafen AG.
Map data copyrighted OpenStreetMap contributors and available from \url{https://www.openstreetmap.org}.
\bibliographystyle{IEEEtran}
\bibliography{format,ref}
\end{document}